\documentclass[11pt]{article}

\usepackage[final]{acl}

\usepackage{times}
\usepackage{latexsym}
\usepackage[T1]{fontenc}
\usepackage[utf8]{inputenc}
\usepackage{microtype}
\usepackage{inconsolata}
\usepackage{graphicx}
\usepackage{hyperref}
\usepackage{url}
\usepackage{booktabs}
\usepackage{amsfonts}
\usepackage{nicefrac}
\usepackage[table]{xcolor}
\usepackage{amsmath}
\usepackage{amssymb}
\usepackage{mathtools}
\usepackage{amsthm}
\usepackage{bm}
\usepackage{multirow}
\usepackage[most]{tcolorbox}
\usepackage{enumitem}
\usepackage{subcaption}
\usepackage{rotating}
\usepackage[normalem]{ulem}
\usepackage{pifont}
\useunder{\uline}{\ul}{}
\usepackage{bbm}
\usepackage{algorithm}
\usepackage{algorithmic}
\usepackage{balance}
\usepackage{array}
\usepackage{makecell}
\usepackage{colortbl}

\newcommand{\mypara}[1]{\vspace{3pt}\noindent{\bf #1}\hspace{5pt}}

\newcommand{\gain}[1]{\textcolor{teal}{\scriptsize(+#1)}}
\newcommand{\loss}[1]{\textcolor{red}{\scriptsize(#1)}}

\def\algname{\textsc{Evo-Harness}}

\title{\algname{}: Context-to-Harness Skill Compilation \\for Self-Evolving Agents}

\author{
\textbf{Tianxin Wei\textsuperscript{1}},
\textbf{Zhan Shi\textsuperscript{2}},
\textbf{Minhua Lin\textsuperscript{3}},
\textbf{Bing He\textsuperscript{2}},
\textbf{Zewen Liu\textsuperscript{4}},
\textbf{Yisi Sang\textsuperscript{2}},
\\
\textbf{Yuanchen Bei\textsuperscript{1}},
\textbf{Xuying Ning\textsuperscript{1}},
\textbf{Jiaru Zou\textsuperscript{1}},
\textbf{Ting-Wei Li\textsuperscript{1}},
\textbf{Xiao Lin\textsuperscript{1}},
\textbf{Yanjun Zhao\textsuperscript{1}},
\\
\textbf{Chi Wang\textsuperscript{5}},
\textbf{Benoit Dumoulin\textsuperscript{2}},
\textbf{Dakuo Wang\textsuperscript{6}},
\textbf{Jingrui He\textsuperscript{1}},
\textbf{Hanqing Lu\textsuperscript{2}}
\\[2mm]
\textsuperscript{1}University of Illinois Urbana-Champaign
\quad
\textsuperscript{2}Amazon
\quad
\textsuperscript{3}The Pennsylvania State University
\\
\textsuperscript{4}Emory University
\quad
\textsuperscript{5}AG2 AI
\quad
\textsuperscript{6}Northeastern University
}

\begin{document}
\maketitle

\begin{abstract}

Learning from experience is critical for developing capable, self-improving large language model (LLM) agents. Existing methods typically extract knowledge from accumulated trajectories via reflection, memory, rules, or skills. However, agents in realistic environments continuously encounter novel tasks, often offering only a one-shot opportunity to improve. These executions yield rich but highly noisy contexts, entangling broadly useful lessons with task-specific artifacts. Critically, prior works rarely validate their effectiveness on complex real-world tasks or isolate the underlying drivers of improvement. To address these gaps, we formulate \emph{online harness learning}, where a frozen agent improves by continually updating a structured harness across sequential tasks. This formulation enables a systematic study of key self-improvement factors through our proposed \emph{Evo-Harness}. At its core, \emph{context-to-harness skill compilation} distills noisy, single-shot executions into reusable skill harnesses for cross-domain and topic-level adaptation. To demonstrate the efficacy of one-shot skill compilation, we evaluate across five realistic benchmarks (TerminalBench2, SWE-bench, CL-Bench, $\tau$-bench, WebArena-Infinity). Our extensive analysis demonstrates the effectiveness of \algname{} and provides a principled understanding of how LLM agents can effectively learn on the fly. Our code is available at
\href{https://github.com/A-EVO-Lab/a-evolve/tree/release/evo-harness}{Link}.

\end{abstract}

\section{Introduction}

Large language model (LLM) agents \cite{achiam2023gpt,team2023gemini,liu2024deepseek} have shown strong performance in complex task-solving settings, such as web interaction \cite{zhou2024webarena}, software engineering \cite{jimenez2024swebench}, tool use \cite{qin2024toolllm}, and long-horizon reasoning \cite{erdogan2025plan,sun2025scaling}. These settings require agents to plan, act, interpret observations, handle feedback, and recover from errors. However, when an agent fails on a task, the failure often remains an isolated event rather than a source of future improvement. Without converting failed executions into reusable lessons, agents continue to make similar mistakes in later tasks.

\begin{figure}
    \centering
    \includegraphics[width=0.7\linewidth]{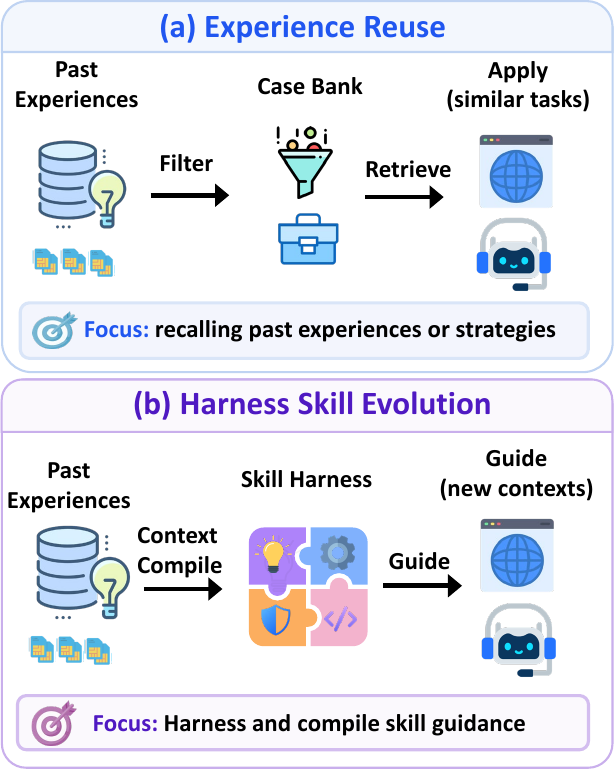}
\caption{
Comparison of experience reuse (retrieving past cases) and harness skill evolution (compiling experiences into reusable guidance).}
\label{fig:teaser}
\vspace{-0.5cm}
\end{figure}

\begin{figure*}
    \centering
    \includegraphics[width=1.03\linewidth]{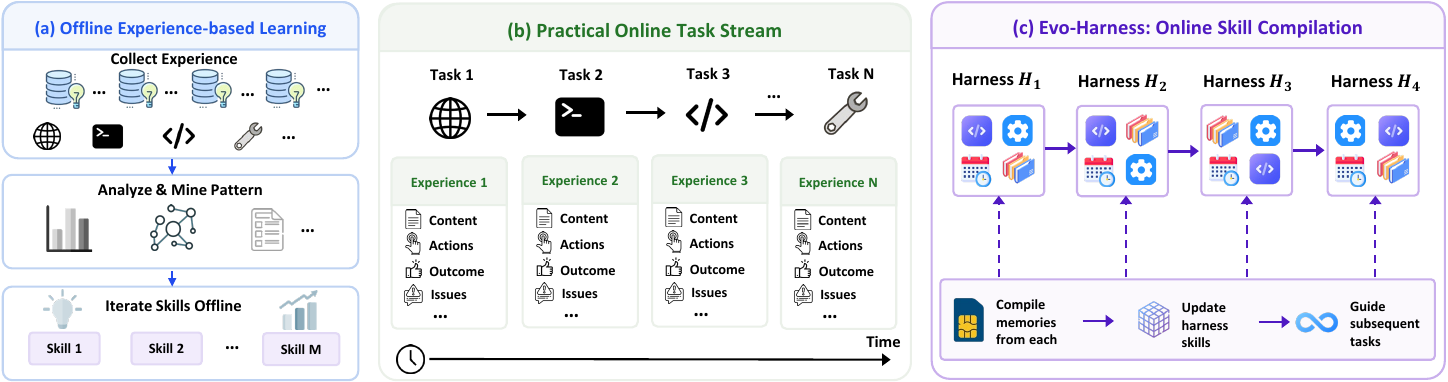}
\caption{
Illustration of the problem setting. (a) Prior methods collect executions and mine reusable patterns offline. (b) In practical online streams, agents encounter tasks sequentially, yielding rich but one-shot execution experiences. (c) \algname{} compiles these online experiences into evolving skill harnesses for future reuse.
}
\label{fig:setting}
\vspace{-0.5cm}
\end{figure*}

Many experience-based methods~\citep{shinn2023reflexion,zhao2024expel} address this issue by learning from accumulated trajectories offline \cite{lee2026meta,ni2026trace2skill}, utilizing media such as reflection \citep{shinn2023reflexion,qiu2026autorefine}, memory \citep{zhong2024memorybank,chhikara2025mem0,xu2025mem}, and skills \citep{zhang2026memskill,xia2026skillrl,jiang2026xskill,yang2026autoskill}. However, in realistic settings, agents continuously encounter novel tasks and contexts, and each completed execution typically offers only a one-shot opportunity to improve future behavior. Furthermore, real executions produce rich but highly noisy contexts, entangling useful lessons with task-specific details, partial failures, and incidental tool traces.

Critically, prior methods often fail to rigorously validate their effectiveness under these realistic constraints, nor do they systematically isolate the underlying drivers of agent improvement. To navigate such a one-shot, complex learning environment, relying on the simple retrieval of past experiences falls short. Instead, agents need to actively compile raw execution contexts into reusable skill guidance for novel tasks, as contrasted in Figure~\ref{fig:teaser}.

Building on this view, we formulate \emph{online harness learning}, where a frozen agent improves by incrementally updating a structured, external harness across sequential tasks. To cope with the continuous influx of novel tasks, this harness acts as a self-evolving learning medium that progressively distills executions into actionable guidance for the future. An overview of this practical setting, compared to standard offline learning, is illustrated in Figure~\ref{fig:setting}.

To operationalize this formulation and enable a systematic study of continuous self-improvement, we introduce \emph{Evo-Harness}. We instantiate the harness within this framework using skills. Skills act as a natural medium for realistic task executions because they can integrate diverse learning signals, such as failures, feedback, constraints, and operational procedures, into structured knowledge that shapes future planning and recovery. At the core of \emph{Evo-Harness} is \emph{context-to-harness skill compilation}, which converts individual execution contexts into reusable skill harness updates. Specifically, our compilation process uses a \emph{solver} to reflect on contexts and propose candidate skills, and an \emph{evolver} to update the existing harness. \emph{General} skills capture transferable cross-task patterns, while \emph{topic} skills capture localized operational knowledge. This dual design preserves critical learning signals while filtering out trajectory-level noise.

We evaluate Evo-Harness across five realistic benchmarks (TerminalBench2, SWE-bench, CL-Bench, $\tau$-bench, and WebArena-Infinity) under a sequential continuous learning setup. Beyond demonstrating performance gains, our primary focus is systematically investigating the core drivers of self-improvement. By analyzing how evolver designs, feedback types, and transfer settings dictate performance, our study treats the skill harness as an interpretable object for understanding how frozen LLM agents learn on the fly.

Our contributions are summarized as follows:
\begin{itemize}[itemsep=0pt,topsep=2pt,leftmargin=*]
    \item \textbf{Online harness learning.}
    We formulate a setting where frozen agents improve over sequential task streams by incrementally updating a reusable external harness instead of model parameters.
    
    \item \textbf{Context-to-harness skill compilation.}
    We introduce a mechanism that compiles single-shot execution contexts into actionable general and topic-level skill updates.
    
    \item \textbf{Systematic analysis of harness evolution.}
    We isolate how evolver designs, feedback types, and transfer settings impact continuous self-improvement. Evaluated across five complex benchmarks, \algname{} consistently outperforms existing experience-based methods.
\end{itemize}

\section{Related Work}\label{sec:related}
In this section, we review existing works on self-evolving LLM agents and external memory or skill harnesses for agent experience.

\paragraph{Evolving and self-improving LLM agents.}
A growing line of work explores how LLM agents self-improve through interaction and feedback. Beyond local reflection methods that iteratively revise outputs based on scalar or verbal feedback~\cite{shinn2023reflexion,madaan2023self}, experience-based approaches attempt to extract reusable knowledge offline. For instance, some works mine accumulated trajectory pools to synthesize natural language insights~\cite{zhao2024expel}, interaction rules~\cite{chen2024automanual}, or transferable directories~\cite{ni2026trace2skill}. Similarly, a parallel track focuses on passively evolving higher-level agent artifacts, such as prompts~\cite{suzgun2025dynamic,lou2026autoharness}, reasoning banks~\cite{ouyang2025reasoningbank}, or memory structures~\cite{agrawal2025gepa,lee2026meta,zhang2025agentic,wei2025evo,lin2026position}. To provide more operational guidance, recent methods have shifted toward skill-centered evolution. This includes building executable code skills~\cite{wang2023voyager}, optimizing skill libraries via policy learning~\cite{xia2026skillrl,zhang2026memskill,ouyang2026skillos}, or curating skill knowledge bases and representations~\cite{yang2026autoskill,qiu2026autorefine,jiang2026xskill,wang2026skillx}. While closely related, these existing frameworks largely treat learned artifacts as passive repositories for free-form retrieval. Critically, prior methods often fail to rigorously validate their effectiveness under realistic online constraints, nor do they systematically isolate the underlying drivers of agent improvement. Our work addresses this gap by compiling single-shot experiences into an active skill harness, providing a structural scaffold to systematically investigate how LLM agents learn on the fly.

\paragraph{Memory and skill harnesses for agent experience.}

Memory provides a fundamental external medium for preserving agent experience. Existing memory systems maintain long-term observation histories~\cite{zhong2024memorybank,yu2025memagent}, organize memories through adaptive structures~\cite{xu2025mem}, and explore methods for updating and retrieving evidence across modalities~\cite{huang2026rethinking,bei2026mem,liu2026omnimem}. However, these approaches largely treat memory as a passive information store. Skills differ fundamentally because they encode operational knowledge. They shift from merely recording what happened to specifying exactly how an agent should act through manuals~\cite{chen2024automanual}, verify operations via organized directories~\cite{ni2026trace2skill}, or recover from failures using skill repositories~\cite{ouyang2026skillos,wang2026skillx,zhang2026mmskills}. This procedural view elevates skills into an active harness. Rather than merely supplying historical context, a harness acts as an external scaffold that structurally shapes future planning, taking forms such as pattern repositories~\cite{ning2026code,qiu2026autorefine}, reasoning banks~\cite{ouyang2025reasoningbank}, and dynamic guidance files~\cite{suzgun2025dynamic,lee2026meta,liu2026adaptive,lin2026harness}. Building on this distinction, our work conceptualizes the skill harness not merely as a performance-enhancing artifact, but as an interpretable medium to systematically study continuous online adaptation. To navigate the high noise of real-world executions, \algname{} compiles single-shot reflections into a structured harness of general and topic skills. While this separation efficiently filters out task-specific artifacts, its primary purpose in our work is to provide a transparent scaffold. This explicitly allows us to systematically isolate and evaluate the core drivers, such as evolver designs and feedback types, that dictate how frozen LLM agents self-improve on the fly.

\section{\algname{}: Context-to-Harness Skill Compilation}
\label{sec:method}

We present \algname{}, a framework for online harness learning. The goal is to convert one-shot execution contexts into a reusable harness that guides future task solving. Unlike experience retrieval, which recalls past cases, \algname{} compiles noisy execution signals into structured guidance that can shape later planning, acting, verification, and recovery.

\begin{figure*}[t]
    \centering
    \includegraphics[width=1.03\textwidth]{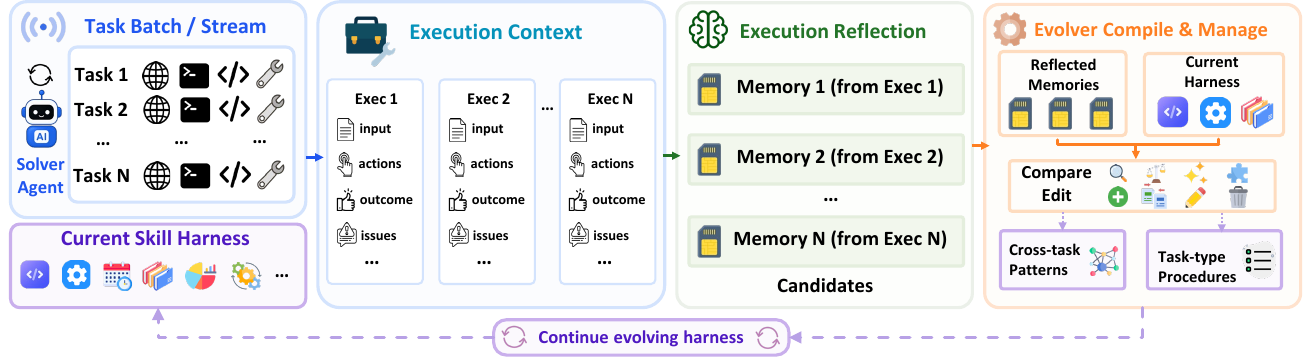}
    \caption{
    Overview of \algname{}. The solver uses the current harness to solve a task batch or stream. Each execution produces a context, including input, action trajectory, outcome, and feedback. Execution reflection converts these contexts into candidate memories. The evolver compares the candidate memories with the current harness and updates the harness with cross-task patterns and task-type procedures for later batches.
    }
    \label{fig:main}
    \vspace{-0.4cm}
\end{figure*}

Figure~\ref{fig:main} gives an overview. \algname{} maintains a current harness that guides the solver before execution, shaping how it plans, acts, verifies, and recovers. After each task batch, execution contexts are reflected into candidate memories and compiled back into the harness as cross-task patterns and task-type procedures, allowing the updated harness to guide later executions.

\subsection{Online Harness Learning Setup}
\label{sec:method-setup}

Let $\{B_1,\ldots,B_K\}$ denote a stream of task batches, where $B_i=\{x_{i,1},\ldots,x_{i,m_i}\}$. Before batch $B_i$, the agent has a current harness $\mathcal{H}_i$, a structured set of reusable guidance entries:
\begin{equation}
    \mathcal{H}_i = \{h_i^1,\ldots,h_i^{n_i}\}.
\end{equation}
The harness is external to the frozen solver and serves as a control scaffold for later executions. Unlike raw trajectory storage, it maintains compact guidance distilled from prior executions and affects future behavior through selection and injection into the solver context.

For each task $x_{i,j}$, the agent selects a compact set of harness entries:
\begin{equation}
    \mathcal{S}_{i,j}
    =
    \mathrm{Select}(x_{i,j}, \mathcal{H}_i; b),
    \qquad
    |\mathcal{S}_{i,j}| \leq b,
\end{equation}
where $b$ is the injection budget. The selected guidance is injected into the task:
\begin{equation}
    \tilde{x}_{i,j}
    =
    \mathrm{Inject}(x_{i,j}, \mathcal{S}_{i,j}).
\end{equation}
A frozen solver $\mathcal{A}$ executes the task:
\begin{equation}
    (\tau_{i,j}, y_{i,j}, f_{i,j})
    =
    \mathcal{A}(\tilde{x}_{i,j}),
\end{equation}
where $\tau_{i,j}$ is the action trajectory, $y_{i,j}$ is the outcome, and $f_{i,j}$ is available feedback, such as verifier results, unit-test outputs, tool diagnostics, or judge feedback. The execution context is:
\begin{equation}
    c_{i,j}
    =
    (x_{i,j}, \tau_{i,j}, y_{i,j}, f_{i,j}).
\end{equation}
Online harness learning updates $\mathcal{H}_i$ into $\mathcal{H}_{i+1}$ from these execution contexts, while keeping the solver parameters fixed.

\subsection{Context-to-Harness Compilation}
\label{sec:method-compilation}

\algname{} uses two complementary stages: \emph{execution reflection} and \emph{harness evolution}. Reflection extracts learning signals from each execution. Evolution integrates those signals into the harness.

% \paragraph{Execution reflection.}
% For each execution context $c_{i,j}$, the solver produces a candidate memory:
% \begin{equation}
%     r_{i,j}
%     =
%     \mathrm{Reflect}(c_{i,j}).
% \end{equation}
% Each candidate memory is represented as:
% \begin{equation}
%     r
%     =
%     (\mathrm{lesson}, \mathrm{trigger}, \mathrm{evidence}, \mathrm{scope\_hint}).
% \end{equation}
% Reflection uses the full task-side context available immediately after execution, including the instruction, trajectory, observations, tool outputs, outcome, and feedback. This enables timely reflection over concrete failures, successful steps, and diagnostic signals before they are abstracted or lost. It does not compare against the current harness, so its output is not a final harness update, but a candidate signal extracted from the completed execution.

\paragraph{Execution reflection.}
For each completed execution context $c_{i,j}$, the solver produces a candidate memory only when the execution fails or receives negative feedback:
\begin{equation}
    r_{i,j}
    =
    \mathrm{Reflect}(c_{i,j}).
\end{equation}
Each candidate memory is represented as:
\begin{equation}
    r
    =
    (\mathrm{lesson}, \mathrm{trigger}, \mathrm{evidence}, \mathrm{scope\_hint}).
\end{equation}
Reflection uses the full task-side context available immediately after execution, including the instruction, trajectory, observations, tool outputs, outcome, and feedback. We focus reflection on failures because they expose the solver's current boundary: incorrect assumptions, missing constraints, ineffective tool use, weak verification, and failed recovery. This makes the resulting candidate memories more targeted for harness improvement, while avoiding unnecessary updates from successful executions that may contain task-specific details. The reflection stage does not compare against the current harness, so its output is not a final harness update, but a failure-grounded candidate signal extracted from the completed execution.

\paragraph{Harness evolution.}
After batch $B_i$, the candidate memories form:
\begin{equation}
    \mathcal{R}_i
    =
    \{r_{i,j}\}_{j=1}^{m_i}.
\end{equation}
The evolver receives the candidate memories and the current harness:
\begin{equation}
    \mathcal{O}_i
    =
    \mathrm{Evolver}(\mathcal{H}_i, \mathcal{R}_i),
\end{equation}
where $\mathcal{O}_i$ is a set of harness edits. Conceptually, the evolver applies:
\begin{equation}
    o
    =
    \pi_{\mathrm{evolve}}(r, \mathcal{H}_i, \mathcal{R}_i),
\end{equation}
where $o \in \{\textsc{Add}, \textsc{Merge}, \textsc{Revise}, \textsc{Skip}\}$. The evolver filters noisy or redundant memories, merges compatible signals, revises existing guidance, and promotes reusable lessons when appropriate. The updated harness is:
\begin{equation}
    \mathcal{H}_{i+1}
    =
    \mathrm{ApplyEdits}(\mathcal{H}_i, \mathcal{O}_i).
\end{equation}
This makes the harness more than a memory store: candidate memories are not appended directly, but compiled into guidance that can affect future behavior.

\subsection{Harness Levels}
\label{sec:method-skill-levels}

The evolved harness contains two levels of reusable guidance. \emph{Cross-task patterns} are compiled by comparing memories across tasks and identifying shared operational structure. \emph{Task-type procedures} preserve more localized guidance for recurring task formats, interfaces, or domains. These levels are not predefined labels for individual executions. They emerge from how the evolver filters, compares, and integrates candidate memories into the harness.

\subsection{Algorithm and Modularity}
\label{sec:method-algorithm}

Algorithm~\ref{alg:context-to-harness} summarizes \algname{}. We use batch-level updates so the evolver can both compile localized task-type procedures and compare memories across tasks to extract cross-task patterns. The same formulation reduces to task-type updating when $|B_i|=1$.

\begin{algorithm}[t]
\caption{\algname{}: Context-to-Harness Skill Compilation}
\label{alg:context-to-harness}
\begin{algorithmic}[1]
\small
\STATE \textbf{Input:} task batches $\{B_1,\ldots,B_K\}$, frozen solver $\mathcal{A}$, initial harness $\mathcal{H}_1$
\FOR{$i=1,\ldots,K$}
    \STATE $\mathcal{R}_i \gets \emptyset$
    \FOR{each task $x_{i,j}\in B_i$}
        \STATE $\mathcal{S}_{i,j}\gets \mathrm{Select}(x_{i,j},\mathcal{H}_i;b)$
        \STATE $\tilde{x}_{i,j}\gets \mathrm{Inject}(x_{i,j},\mathcal{S}_{i,j})$
        \STATE $(\tau_{i,j},y_{i,j},f_{i,j})\gets \mathcal{A}(\tilde{x}_{i,j})$
        \STATE $c_{i,j}\gets (x_{i,j},\tau_{i,j},y_{i,j},f_{i,j})$
        \STATE $r_{i,j}\gets \mathrm{Reflect}(c_{i,j})$
        \STATE $\mathcal{R}_i\gets \mathcal{R}_i\cup\{r_{i,j}\}$
    \ENDFOR
    \STATE $\mathcal{O}_i^{\mathrm{type}}\gets \mathrm{CompileTaskType}(\mathcal{H}_i,\mathcal{R}_i)$
    \STATE $\mathcal{O}_i^{\mathrm{cross}}\gets \mathrm{CompileCrossTask}(\mathcal{H}_i,\mathcal{R}_i)$
    \STATE $\mathcal{O}_i\gets \mathcal{O}_i^{\mathrm{type}}\cup \mathcal{O}_i^{\mathrm{cross}}$
    \STATE $\mathcal{H}_{i+1}\gets \mathrm{ApplyEdits}(\mathcal{H}_i,\mathcal{O}_i)$
\ENDFOR
\STATE \textbf{Output:} evolved harness $\mathcal{H}_{K+1}$ and task results
\end{algorithmic}
\end{algorithm}

Here, $\mathrm{CompileTaskType}$ preserves localized procedures, while $\mathrm{CompileCrossTask}$ extracts shared patterns across executions. The modular design is intentional: by decoupling the solver, evolver, feedback source, and initial harness state, \algname{} provides a controlled framework for systematically studying what drives online harness improvement.
\section{Experiments}\label{sec:exp}
\label{sec:experiments}

We conduct experiments to evaluate \algname{} and to understand when online harness evolution improves LLM agents:
\begin{itemize}[leftmargin=12pt,noitemsep,topsep=2pt]
    \item \textbf{RQ1: Overall effectiveness.}
    Does \algname{} improve frozen agents across diverse benchmarks, and which task categories benefit most?
    \item \textbf{RQ2: Harness content and components.}
    What types of guidance are produced by the evolved harness, and how do different harness components contribute?
    \item \textbf{RQ3: Online updating and transfer.}
    How do evolved harnesses behave under train-split transfer, online updating, and solver-evolver pairing?
    \item \textbf{RQ4: Feedback grounding.}
    How does the source and granularity of feedback affect harness evolution?
\end{itemize}

\subsection{Setup}\label{sec:exp-setup}

\mypara{Experimental setup.}
We evaluate \algname{} on five benchmarks spanning diverse agent capabilities: WebArena-Infinity~\citep{zhou2026wainf} for web navigation, TerminalBench2~\citep{merrill2026terminalbenchbenchmarkingagentshard} for command-line tasks, SWE-bench Lite~\citep{jimenez2024swebench} for software engineering, CL-Bench~\citep{dou2026cl} for adaptive reasoning, and TAU-Bench~\citep{yao2024tau} for tool-use tasks. We use Claude Opus 4.6 \cite{claude_opus_46} as the default solver and evolver, and additionally evaluate with Claude Opus 4.7 \cite{claude_opus_47}, Claude Opus 4.5 \cite{claude_opus_45}, Kimi-K2.5 \cite{team2026kimi}, and GPT-OSS \cite{agarwal2025gpt} when studying model ablations. We report success rates in percent, where higher is better. Further benchmark and implementation details are provided in Appendix~\ref{app:benchmarks} and \ref{app:details}.

\mypara{Baselines.}
We compare with representative experience-, memory-, and skill-based adaptation methods under the same task split and evaluation protocol. AWM~\citep{wang2024agent} reuses agent working memory, Dynamic Cheatsheet~\citep{suzgun2025dynamic} maintains an adaptive external cheatsheet, Evo-Memory~\citep{wei2025evo} evolves reusable memories from executions, ACE~\citep{zhang2025agentic} optimizes agent context, and XSkill~\citep{jiang2026xskill} separates experience and skills for continual reuse. DC-Cu and DC-RS are two Dynamic Cheatsheet variants.

% \tingwei{just curious, since we are in an sequential setting (and previous learned harness will impact latter tasks), does the ordering of the tasks in each dataset during eval matter? do we have a specific order or they are just random?}

\begin{figure}
    \centering
    \includegraphics[width=\linewidth]{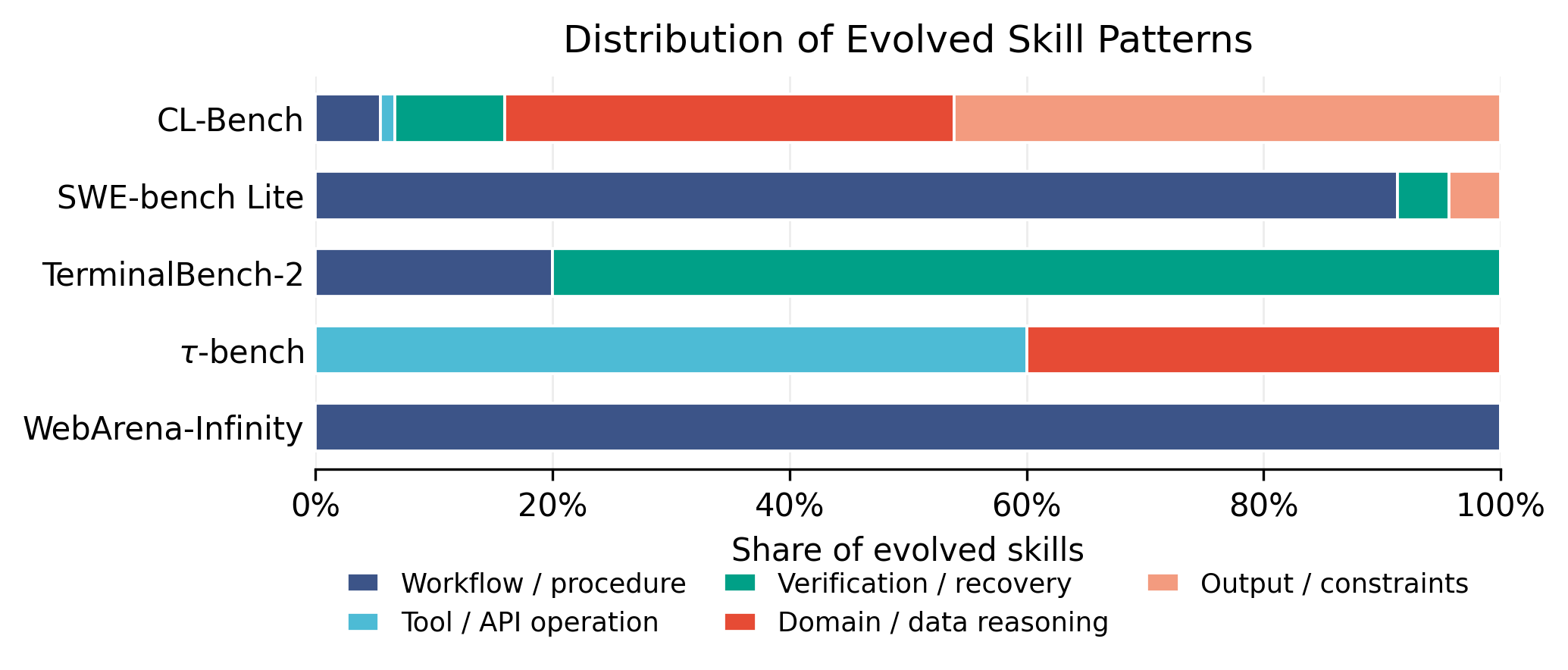}
    \caption{Distribution of evolved skill patterns across benchmarks. The learned skills exhibit benchmark-specific characteristics, ranging from workflow and tool-use procedures to verification, domain reasoning, and output-constraint handling.}    
    \label{fig:skill-taxonomy}
\end{figure}

\begin{table*}[t]
\caption{
Main results using Claude Opus 4.6 as the solver. Scores denote success rates (\%), and higher values indicate better performance. \algname{} achieves the best results across all five benchmarks.
}
\centering
\label{tab:main}
\small
\begin{tabular}{lccccc}
\toprule
\textbf{Method} & \textbf{CL-Bench} & \textbf{TerminalBench-2} & \textbf{SWE-bench Lite} & \textbf{$\tau$-bench} & \textbf{WebArena-Infinity} \\
\midrule
No Evolve & 29.54 & 62.92 & 63.67 & 72.73 & 72.50 \\
\midrule
AWM & 28.91 & 62.92 & 62.67 & 70.91 & 70.00 \\
DC-Cu & 29.23 & 60.67 & 63.00 & 72.12 & 71.25 \\
DC-RS & 29.07 & 61.80 & 61.67 & 70.91 & 70.00 \\
Evo-Memory & 29.38 & 64.04 & 64.00 & 72.12 & 72.50 \\
ACE & 29.70 & 61.80 & 63.67 & 72.73 & 71.25 \\
XSkill & 31.44 & 66.29 & 64.67 & 73.94 & 73.75 \\
\midrule
\algname{} & \textbf{34.02} & \textbf{73.03} & \textbf{67.00} & \textbf{76.97} & \textbf{76.25} \\
\bottomrule
\end{tabular}
\label{tab:main-opus46}
\end{table*}

\subsection{Overall Effectiveness (RQ1)}
\label{sec:exp-main}

Table~\ref{tab:main} reports the main results with Claude Opus 4.6 as the solver. 
\textbf{\algname{} improves over both No-Evolve and prior experience-based baselines across all five benchmarks.}
The gains are especially large on TerminalBench-2, where agents need to inspect files, execute commands, interpret errors, and recover from failed operations. This suggests that harness evolution is particularly useful when tasks expose reusable procedural structure.

Compared with the strongest baseline, XSkill, \algname{} still improves on every benchmark. 
\textbf{External experience reuse is not automatically beneficial.}
Several baselines underperform No-Evolve on some benchmarks, suggesting that retrieved or evolved guidance can be noisy, overly specific, or mismatched to the solver. This supports our motivation for studying not only whether self-evolution helps, but also what factors make online harness updates effective.

\subsection{Category and Model Generality (RQ1)}
\label{sec:exp-category-model}

Table~\ref{tab:clbench-category} provides a representative category-level view on CL-Bench across five solver models. Since CL-Bench covers diverse reasoning and execution categories, it allows us to examine where harness evolution is most effective and how the effect varies across solver capabilities.
\textbf{\algname{} improves the overall score for every solver model, with larger gains on stronger Claude models than on open-weight alternatives.}
Opus 4.7, Opus 4.6, and Opus 4.5 obtain gains of $+3.7$, $+4.5$, and $+3.8$, respectively, while Kimi-K2.5 and GPT-OSS improve by $+1.1$ and $+0.8$. This suggests that stronger solvers may be better able to interpret, follow, and benefit from evolved harness guidance.

Across categories, the most consistent improvements appear on Procedural Task Execution (PTE), where every model improves and Opus 4.6 and Opus 4.5 each gain $+9.8$ points. 
\textbf{Harness evolution is most reliable when tasks expose reusable operational structure.}
In contrast, Empirical Discovery \& Simulation (EDS) shows less stable gains, with drops for Opus 4.6 and GPT-OSS. This suggests that more exploratory tasks may be more sensitive to overly specific guidance.

\begin{table*}[t]
\centering
\caption{CL-Bench category results across different models: pass rate (\%) across five models. \textbf{Bold} = best per category-model pair. $\Delta$ shows absolute improvement of \algname{} over the No-Evolve baseline. DKR = Domain Knowledge Reasoning, EDS = Empirical Discovery \& Simulation, PTE = Procedural Task Execution, RSA = Rule System Application.}
\label{tab:clbench-category}
\vspace{3pt}
\small
\setlength{\tabcolsep}{4pt}
\begin{tabular}{ll|cc|cc|cc|cc|cc}
\toprule
& & \multicolumn{2}{c|}{\textbf{DKR}} & \multicolumn{2}{c|}{\textbf{EDS}} & \multicolumn{2}{c|}{\textbf{PTE}} & \multicolumn{2}{c|}{\textbf{RSA}} & \multicolumn{2}{c}{\textbf{All}} \\
\textbf{Model} & \textbf{Method} & Rate & $\Delta$ & Rate & $\Delta$ & Rate & $\Delta$ & Rate & $\Delta$ & Rate & $\Delta$ \\
\midrule
\multirow{2}{*}{Opus 4.7}
& No-Evolve & 32.4 & --- & \textbf{19.1} & --- & 35.5 & --- & 29.0 & --- & 30.8 & --- \\
& \algname{} & \textbf{34.1} & \gain{1.7} & 19.6 & \gain{0.5} & \textbf{42.7} & \gain{7.2} & \textbf{32.7} & \gain{3.7} & \textbf{34.5} & \gain{3.7} \\
\midrule
\multirow{2}{*}{Opus 4.6}
& No-Evolve & 31.4 & --- & \textbf{20.6} & --- & 34.0 & --- & 26.9 & --- & 29.5 & --- \\
& \algname{} & \textbf{35.6} & \gain{4.2} & 18.1 & \loss{-2.5} & \textbf{43.8} & \gain{9.8} & \textbf{30.4} & \gain{3.5} & \textbf{34.0} & \gain{4.5} \\
\midrule
\multirow{2}{*}{Opus 4.5}
& No-Evolve & 31.4 & --- & 18.1 & --- & 27.8 & --- & 25.3 & --- & 27.3 & --- \\
& \algname{} & \textbf{32.6} & \gain{1.2} & \textbf{19.1} & \gain{1.0} & \textbf{37.6} & \gain{9.8} & \textbf{28.3} & \gain{3.0} & \textbf{31.1} & \gain{3.8} \\
\midrule
\multirow{2}{*}{Kimi-K2.5}
& No-Evolve & 25.6 & --- & 12.1 & --- & 25.1 & --- & \textbf{21.6} & --- & 22.9 & --- \\
& \algname{} & \textbf{25.9} & \gain{0.3} & \textbf{13.1} & \gain{1.0} & \textbf{29.3} & \gain{4.2} & 20.8 & \loss{-0.7} & \textbf{23.9} & \gain{1.1} \\
\midrule
\multirow{2}{*}{GPT-OSS}
& No-Evolve & 17.6 & --- & \textbf{11.6} & --- & 17.6 & --- & 16.4 & --- & 16.6 & --- \\
& \algname{} & \textbf{17.9} & \gain{0.3} & 10.6 & \loss{-1.0} & \textbf{18.5} & \gain{0.8} & \textbf{18.6} & \gain{2.1} & \textbf{17.5} & \gain{0.8} \\
\bottomrule
\end{tabular}
\vspace{-5pt}
\end{table*}

\begin{table}[t]
\centering
\caption{Ablation study on CL-Bench and SWE-bench Lite using Claude Opus 4.6 as the solver. Scores denote pass rates (\%).}
\label{tab:ablation-cl-swe}
\vspace{3pt}
\small
\begin{tabular}{lcc}
\toprule
\textbf{Method} & \textbf{CL-Bench} & \textbf{SWE-bench Lite} \\
\midrule
No Evolve & 29.54 & 63.67 \\ \midrule
No Propose & 33.28 & 65.33 \\
General Only & 30.28 & 66.67 \\
Topic Only & 33.70 & 64.33 \\
\midrule
\algname{} & \textbf{34.02} & \textbf{67.00} \\
\bottomrule
\end{tabular}
\vspace{-5pt}
\end{table}

\subsection{Harness Content and Components (RQ2)}
\label{sec:exp-harness-analysis}

% Figure~\ref{fig:skill-taxonomy} analyzes the content of the evolved harness across benchmarks. 
% \textbf{The evolved harness reflects benchmark-specific operational demands rather than accumulating generic guidance.} WebArena-Infinity is dominated by workflow and procedure skills, which is consistent with interface-driven tasks that require multi-step navigation and state manipulation. TerminalBench-2 contains a large share of verification and recovery skills, reflecting the need to inspect command outputs, diagnose failures, and validate intermediate states. $\tau$-bench contains a mixture of tool/API operation and domain/data reasoning skills, matching its structured tool-use setting. CL-Bench shows a broader distribution, including domain reasoning and output-constraint handling, which is consistent with its adaptive reasoning tasks. These patterns support our view that context-to-harness compilation adapts the harness to the operational structure and failure modes of each environment.

Figure~\ref{fig:skill-taxonomy} analyzes the content of the evolved harness across benchmarks. 
\textbf{The evolved harness reflects benchmark-specific operational demands rather than accumulating generic guidance.}
WebArena-Infinity mainly yields workflow and procedure skills, TerminalBench-2 emphasizes verification and recovery, $\tau$-bench mixes tool/API operation with domain reasoning, and CL-Bench produces a broader distribution including domain reasoning and output-constraint handling. These patterns suggest that context-to-harness compilation adapts the harness to each environment's operational structure and failure modes.

Table~\ref{tab:ablation-cl-swe} studies which parts of the harness update process contribute to performance. We focus this and the following diagnostic analyses mainly on CL-Bench and SWE-bench Lite because they stress different forms of harness guidance: CL-Bench contains heterogeneous adaptive reasoning tasks where localized task-type guidance can matter, while SWE-bench Lite emphasizes repository-level debugging and verification procedures where broader reusable guidance is expected to help. \textsc{No Propose} removes solver-side proposal of candidate memories, \textsc{General Only} keeps only cross-task guidance, and \textsc{Topic Only} keeps only localized task-type guidance.

% \textbf{Both broad cross-task guidance and localized task-type guidance are useful, but their importance depends on the task setting.}
% The full \algname{} setting achieves the best performance on both benchmarks, reaching $34.02$ on CL-Bench and $67.00$ on SWE-bench Lite. On CL-Bench, Topic Only reaches $33.70$, close to the full method, while General Only reaches $30.28$, suggesting that CL-Bench benefits from localized operational knowledge associated with recurring task types or contexts. On SWE-bench Lite, the pattern is reversed: General Only reaches $66.67$, close to the full method, while Topic Only reaches $64.33$, suggesting that SWE-bench Lite benefits more from broad procedures such as repository inspection, debugging, verification, and test-driven repair. No Propose improves over No Evolve on both benchmarks but remains below the full method, indicating that proposal and harness-level compilation are complementary.

\textbf{Both broad cross-task guidance and localized task-type guidance are useful, but their importance depends on the task setting.}
The full \algname{} setting performs best on both benchmarks. On CL-Bench, Topic Only reaches $33.70$, close to the full result of $34.02$, while General Only drops to $30.28$, suggesting that localized task-type guidance is more useful for its heterogeneous reasoning tasks. On SWE-bench Lite, General Only reaches $66.67$, close to the full result of $67.00$, while Topic Only reaches $64.33$, indicating that broad debugging, verification, and repository-inspection procedures matter more. No Propose improves over No Evolve but remains below the full method, suggesting that proposal and harness-level compilation are complementary.

\begin{figure}
    \centering
    \includegraphics[width=0.7\linewidth]{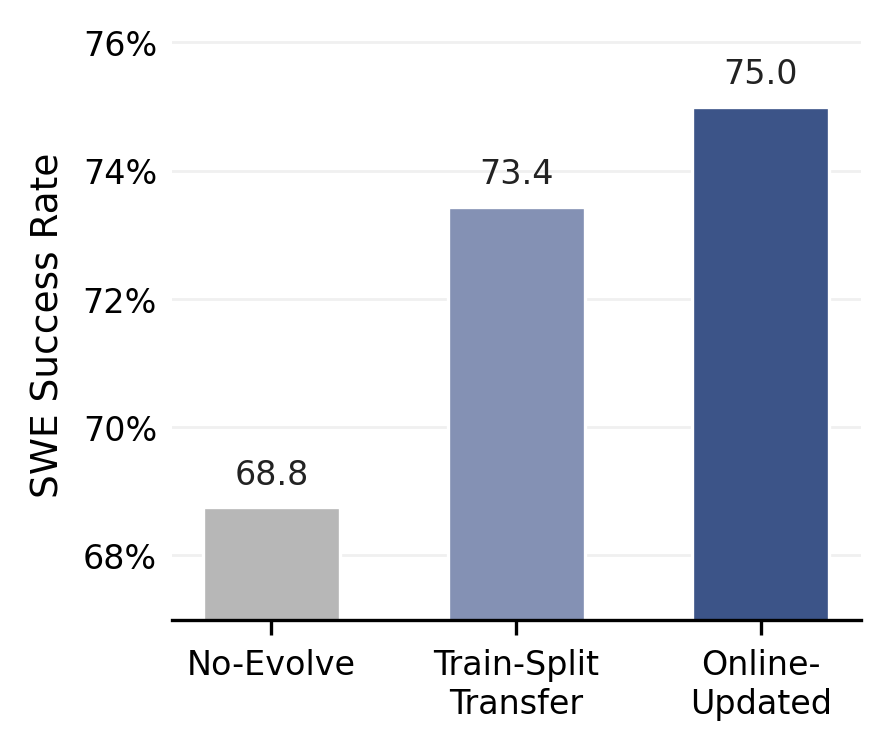}
\caption{SWE-bench Lite success rates under different evolution settings. 
\textsc{No-Evolve} uses no harness updates; \textsc{Train-Split Transfer} evaluates skills evolved by Claude Sonnet on the training split; and \textsc{Online-Updated} updates skills during the task stream. }
    \label{fig:swe-evolution-setting}
    \vspace{-0.5cm}
\end{figure}

% ===========================================================================
\subsection{Transfer and Adaptability (RQ3)}
\label{sec:exp-transfer}
% ===========================================================================

Figure~\ref{fig:swe-evolution-setting} compares three skill-update settings. \textsc{No-Evolve} uses no harness update. \textsc{Train-Split Transfer} uses skills evolved by Claude Sonnet 4.5 on the training split and evaluates them with an Opus 4.7 solver on the test split. \textsc{Online-Updated} updates the harness directly with the same models during the test task stream. 
\textbf{Skills learned on a training split can transfer across both tasks and models, while online updating performs best.}
No-Evolve reaches $68.8$, Train-Split Transfer reaches $73.4$, and Online-Updated reaches $75.0$. The gain from Train-Split Transfer shows that skills evolved by a smaller or different model can still provide useful guidance for a stronger solver on unseen tasks. Online-Updated further improves over Train-Split Transfer by $+1.6$ points, suggesting that in-situ harness updates better adapt to the local distribution and failure modes of the test stream.

\begin{figure}[t]
  \centering
  \begin{subfigure}{0.48\linewidth}
    \centering
    \includegraphics[width=\linewidth]{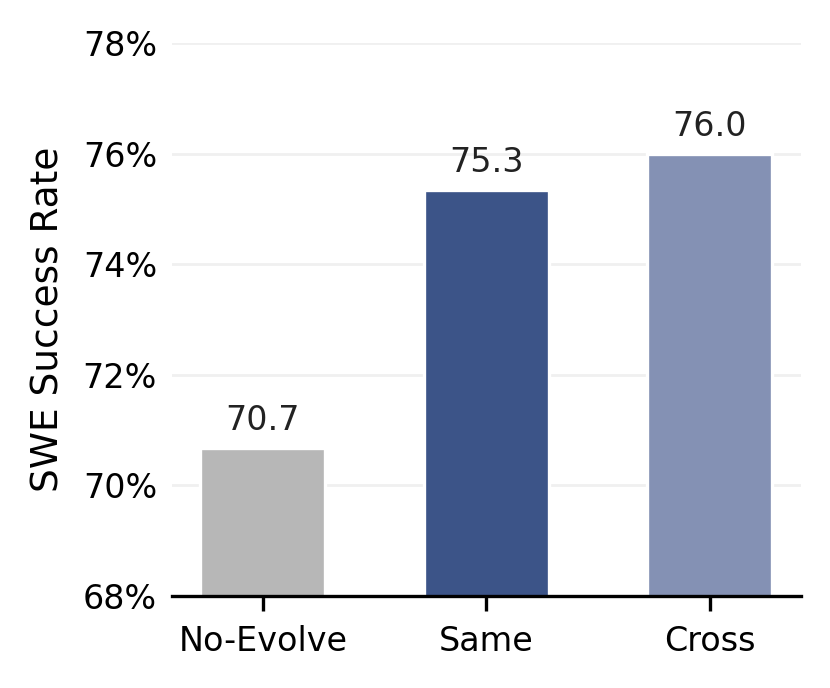}
    \caption{Opus 4.7 Solver}
    \label{fig:swe-curator-opus}
  \end{subfigure}
  \hfill
  \begin{subfigure}{0.48\linewidth}
    \centering
    \includegraphics[width=\linewidth]{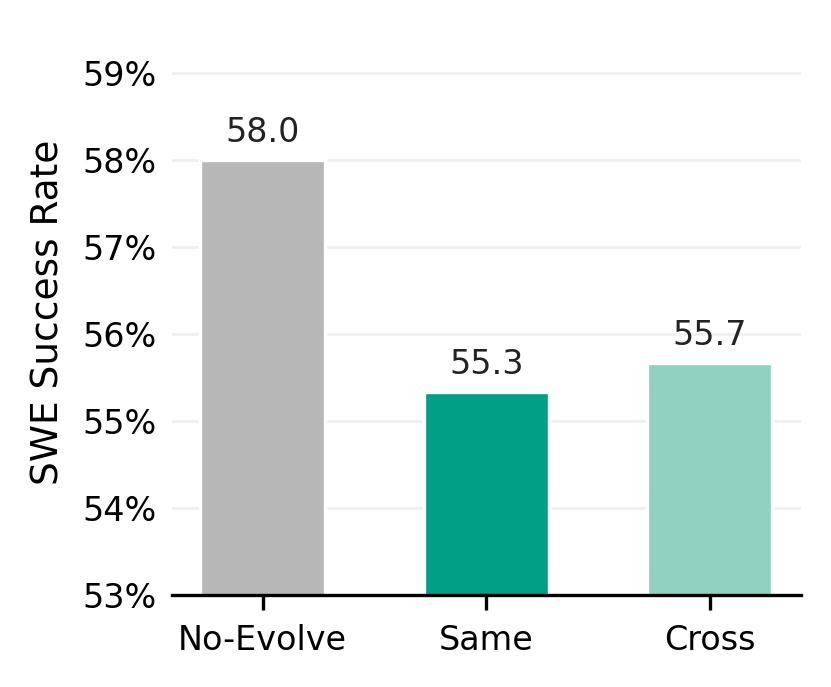}
    \caption{Sonnet 4.5 Solver}
    \label{fig:swe-curator-sonnet}
  \end{subfigure}
\caption{Effect of solver-evolver pairing on SWE-bench Lite. 
\textsc{Same} denotes using the same model for solving and skill evolution, while \textsc{Cross} denotes using different models for the solver and evolver. }
  \label{fig:swe-curator-solver}
  \vspace{-0.4cm}
\end{figure}

Figure~\ref{fig:swe-curator-solver} examines how solver-evolver pairing affects harness evolution. We compare two solver settings: an Opus 4.7 solver and a Sonnet 4.5 solver. In each setting, \textsc{Same} uses the same model as both solver and evolver, while \textsc{Cross} uses the other model as the evolver. Specifically, for the Opus solver, \textsc{Same} uses Opus to evolve the harness and \textsc{Cross} uses Sonnet to evolve the harness; for the Sonnet solver, \textsc{Same} uses Sonnet as the evolver and \textsc{Cross} uses Opus as the evolver.
\textbf{Using the same model for solving and evolution is not always optimal; cross-model evolution can help, but only when the solver is capable of using the evolved guidance.}
For the Opus solver, Cross reaches $76.0$, slightly higher than Same at $75.3$ and clearly above No-Evolve at $70.7$. This suggests that a different evolver can produce guidance that is still useful, or even more useful, when the downstream solver has sufficient capability to interpret and apply it. For the Sonnet solver, however, Same and Cross reach $55.3$ and $55.7$, both below No-Evolve at $58.0$. Thus, Sonnet-evolved skills can help Opus, but do not necessarily help Sonnet itself. This contrast suggests that harness transfer depends not only on the skill artifact but also on the base solver’s capability: if the solver cannot reliably follow, contextualize, or adapt the guidance, evolution may fail to improve performance.

% ===========================================================================
\subsection{Feedback Grounding (RQ4)}
\label{sec:exp-feedback}

% ===========================================================================

Table~\ref{tab:feedback-level-ablation} evaluates how feedback signals affect harness evolution on CL-Bench and SWE-bench Lite. We compare three feedback settings. \textsc{Self-Generated} asks the LLM itself to judge whether the execution succeeded and what should be learned. \textsc{Minimal} provides only the environment-level success or failure signal, such as whether a command, verifier, or test passed. \textsc{Standard} provides grounded diagnostic feedback, including failure messages, error traces, verifier outputs, or test failures when available.

\textbf{Grounded feedback is important for reliable harness evolution.}
Self-Generated feedback underperforms No-Evolve on both benchmarks, decreasing CL-Bench from $29.54$ to $27.96$ and SWE-bench Lite from $63.67$ to $61.67$. This suggests that LLM-generated self-judgment can introduce misleading updates when it is not grounded in external execution evidence. In contrast, Minimal and Standard feedback both use environment-provided signals and are more reliable overall.

\textbf{Standard feedback is generally preferable, but the useful granularity of feedback depends on the benchmark.}
On CL-Bench, Standard feedback performs best, reaching $34.02$, while Minimal feedback reaches only $29.86$. This suggests that richer diagnostics are useful for adaptive reasoning tasks, where understanding the reason for failure can help compile better harness guidance. On SWE-bench Lite, however, Minimal feedback reaches $67.33$, slightly above Standard feedback at $67.00$. One possible explanation is that detailed error messages can sometimes make the evolved harness overly tied to task-specific failures, while sparse pass/fail signals encourage more generalizable guidance.

\begin{table}[t]
\centering
\caption{Impact of feedback level on CL-Bench and SWE-bench Lite using Claude Opus 4.6 as the solver. Scores denote pass rates (\%).}
\label{tab:feedback-level-ablation}
\vspace{3pt}
\small
\begin{tabular}{lcc}
\toprule
\textbf{Feedback Level} & \textbf{CL-Bench} & \textbf{SWE-bench Lite} \\
\midrule
No Evolve & 29.54 & 63.67 \\
Self-Generated & 27.96 & 61.67 \\
Minimal & 29.86 & \textbf{67.33} \\
Standard & \textbf{34.02} & 67.00 \\
\bottomrule
\end{tabular}
\vspace{-5pt}
\end{table}

\section{Conclusion}\label{sec:con}

We formulated \emph{online harness learning} as a setting for studying how frozen LLM agents can improve from sequential, one-shot executions. Through \algname{}, we instantiated this setting with context-to-harness skill compilation, which turns noisy execution contexts into a reusable external harness for future task solving. Beyond proposing a pipeline, our goal is to use the harness as an analytical lens for understanding online self-improvement. This perspective moves self-evolving agent research beyond aggregate success rates toward a more principled study of how agents produce, organize, transfer, and apply reusable experience in realistic task streams.

\section*{Limitations}

Our study focuses on single-agent LLM systems in text-based, tool-use, web, command-line, and software-engineering environments. We do not evaluate embodied agents or multi-agent systems, where the harness may involve additional forms of interaction, coordination, and environmental feedback. We also instantiate the harness with natural-language guidance; other harness formats, such as executable code skills or structured programs, are not covered in this study.

\bibliography{main}
\clearpage
\newpage
\appendix
\section*{Appendix}

\tcbset{
  evoCase/.style={
    enhanced,
    breakable,
    arc=1mm,
    boxrule=0.45pt,
    left=4pt,
    right=4pt,
    top=4pt,
    bottom=4pt,
    fonttitle=\bfseries\small,
    coltitle=black
  },
  skillCard/.style={
    enhanced,
    arc=1mm,
    boxrule=0.35pt,
    left=4pt,
    right=4pt,
    top=3pt,
    bottom=3pt,
    fonttitle=\bfseries\footnotesize,
    coltitle=black
  }
}

\section{Benchmark Details}\label{app:benchmarks}

Table~\ref{tab:bench-stats} summarizes the benchmark settings. We use the same
online protocol across benchmarks: tasks are processed as a stream, the solver
uses the current harness, and the evolver updates the harness after observing
task outcomes.

\begin{table}[h]
\centering
\caption{Benchmark statistics and evaluation signals.}
\label{tab:bench-stats}
\small
\setlength{\tabcolsep}{3pt}
\scalebox{0.78}{
\begin{tabular}{lrll}
\toprule
\textbf{Benchmark} & \textbf{\#Tasks} & \textbf{Domain} & \textbf{Evaluation signal} \\
\midrule
WebArena Inf. & 80 & Web navigation & State verifier \\
Terminal-Bench & 89 & CLI / scripting & Docker verifier \\
SWE-bench Lite & 300 & Software eng. & Unit tests \\
CL-Bench & 1899 & Adaptive reasoning & Rubric judge \\
TAU-Bench & 165 & Tool use & State verifier \\
\bottomrule
\end{tabular}}
\end{table}

\paragraph{WebArena Infinity.}
We use 80 hard tasks from the email-centered ``superhuman-general'' setting and
related web applications. Tasks require stateful browser interaction, such as
creating labels, changing settings, using email filters, or coordinating
calendar/email information. A programmatic verifier checks the final application
state. This benchmark is useful for studying whether skills capture UI
procedures.

\paragraph{Terminal-Bench v2.}
We evaluate 89 tasks across software engineering, data processing, scientific
computing, security, system administration, and related command-line domains.
The solver works in an isolated shell environment and submits final artifacts or
answers inside a task-specific Docker image. Final success is determined by the
benchmark's executable verifier, typically a hidden or mounted test script run
inside the container. 

\paragraph{SWE-bench Lite.}
We use 300 real GitHub issues from 12 Python repositories. The solver edits code
and the final patch is evaluated by repository tests.

\paragraph{CL-Bench.}
CL-Bench contains 1899 adaptive reasoning tasks across Domain Knowledge
Reasoning, Empirical Discovery \& Simulation, Procedural Task Execution, and
Rule System Application. Each task is graded with rubric criteria.

\paragraph{TAU-Bench.}
TAU-Bench contains 165 customer-service tool-use tasks across airline and
retail domains. Evaluation checks the final database state.

\section{Harness Artifacts Examples}\label{app:examples}

This section shows representative learned artifacts without enumerating the
learned inventory. For task-level comparison, we align the same task identifiers
between the no-evolve and evolved runs. We only call a case a direct improvement
when the baseline run fails and the evolved run succeeds. This is not a
single-skill causal ablation: several skills may be injected together, and some
benchmarks do not record injected skills in every result file. The examples
therefore use a structured failure/success format: task, verifier signal,
trajectory summary, learned skill content, and a short takeaway. 

\begin{tcolorbox}[evoCase, colback=blue!2, colframe=blue!45,
title={WebArena: verified setting}]
\small
\begin{itemize}[leftmargin=1.1em, itemsep=1pt, topsep=1pt]
\item \textbf{Task:} \texttt{task\_h10}; create ``Legal'' and disable
``Support Ticket''.
\item \textbf{Baseline:} verifier found ``Support Ticket'' still enabled,
despite the trajectory claiming both subgoals were complete.
\item \textbf{Evolved:} verifier passed; the trajectory returned to the Auto
Labels list and checked the existing label state.
\item \textbf{Learned skill:} toggle an existing auto label off; do not delete
it; verify creation and disabling independently.
\end{itemize}
\tcblower
\textbf{Core point.} The harness learned a concrete UI distinction:
create-new-label and disable-existing-label are separate operations.
\end{tcolorbox}

\begin{tcolorbox}[evoCase, colback=blue!2, colframe=blue!45,
title={WebArena: grounded recipients}]
\small
\begin{itemize}[leftmargin=1.1em, itemsep=1pt, topsep=1pt]
\item \textbf{Task:} \texttt{task\_h65}; reply to Kevin Zhao with CTO and VP
Engineering on CC.
\item \textbf{Baseline:} sent to plausible names but wrong domains
(\texttt{acme.com} instead of \texttt{acmecorp.com}).
\item \textbf{Evolved:} verifier passed with the required role-derived
recipients in CC.
\item \textbf{Learned skill:} prefer authoritative role sources and preserve
exact email addresses across apps.
\end{itemize}
\tcblower
\textbf{Core point.} The harness learned identity grounding, not merely email
composition.
\end{tcolorbox}

\begin{tcolorbox}[evoCase, colback=orange!2, colframe=orange!45,
title={Terminal-Bench: executable recovery}]
\small
\begin{itemize}[leftmargin=1.1em, itemsep=1pt, topsep=1pt]
\item \textbf{Task:} \texttt{adaptive-rejection-sampler}; pass the Docker
verifier for an R sampler.
\item \textbf{Baseline:} formal tests failed after repeated reruns of the same
R test command.
\item \textbf{Evolved:} passed after diagnosing timeout, missing dependency,
formal-test, and sorted-vector failures.
\item \textbf{Learned skill:} after repeated similar failures, stop, shrink the
case, build a minimal reproducer, then add complexity back.
\end{itemize}
\tcblower
\textbf{Core point.} The harness learned a recovery policy for executable
feedback loops.
\end{tcolorbox}

\begin{tcolorbox}[evoCase, colback=green!2, colframe=green!45,
title={SWE-bench: test-grounded repair}]
\small
\begin{itemize}[leftmargin=1.1em, itemsep=1pt, topsep=1pt]
\item \textbf{Task:} \texttt{django\_\_django-11133}; fix Django response
handling without regressing existing behavior.
\item \textbf{Baseline:} timed out after source search and failed test attempts.
\item \textbf{Evolved:} passed \texttt{FAIL\_TO\_PASS: 1/1} and
\texttt{PASS\_TO\_PASS: 64/64}.
\item \textbf{Learned skill:} read complete failing-test assertions, match
exact expected behavior, then run targeted tests before submission.
\end{itemize}
\tcblower
\textbf{Core point.} The harness learned to anchor patches in executable test
expectations.
\end{tcolorbox}

\begin{tcolorbox}[evoCase, colback=purple!2, colframe=purple!45,
title={TAU-Bench: complete tool execution}]
\small
\begin{itemize}[leftmargin=1.1em, itemsep=1pt, topsep=1pt]
\item \textbf{Task:} \texttt{retail\_task\_5}; exchange multiple items from one
order.
\item \textbf{Baseline:} database verifier failed, consistent with partial or
wrong-tool execution.
\item \textbf{Evolved:} verifier passed after retrieving the order and applying
all requested item changes.
\item \textbf{Learned skill:} enumerate every requested operation, choose the
tool by order status, and verify the response covers the full list.
\end{itemize}
\tcblower
\textbf{Core point.} The harness learned complete tool-call execution.
\end{tcolorbox}

\section{Which Tasks Are Easier to Improve?}\label{app:improvement-hotspots}

The most informative pattern is not the total number of fail-to-pass cases, but
the kind of task that turns around after harness evolution.
Table~\ref{tab:improvement-hotspots} groups matched fail-to-pass cases by their
observable failure mode and gives representative task identifiers from the
runs in \texttt{outputs}.

\begin{table*}[t]
\centering
\caption{Improvement hotspots from matched baseline-fail/evolved-pass cases.}
\label{tab:improvement-hotspots}
\small
\setlength{\tabcolsep}{3.5pt}
\begin{tabular}{p{2.0cm}p{3.2cm}p{4.3cm}p{4.0cm}}
\toprule
\textbf{Benchmark} & \textbf{Easier-to-improve tasks} &
\textbf{Representative cases} & \textbf{Why skills help} \\
\midrule
WebArena &
Stateful UI tasks with exact recipients, labels, settings, and mail state. &
\texttt{h10}: disable an existing auto label;
\texttt{h65}: preserve corporate email domains;
\texttt{h73}: archive all qualifying read-important emails. &
The verifier checks final app state, so procedural memory about where to click
must be coupled with exact state confirmation. \\
\midrule
Terminal-Bench &
Tasks with executable feedback loops and recoverable failures. &
\texttt{adaptive-rejection-sampler}, \texttt{dna-insert},
\texttt{build-pmars}, \texttt{regex-chess}. &
The solver can observe Docker-test failures; process skills help it pivot,
reduce scope, and verify artifacts before submission. \\
\midrule
SWE-bench Lite &
Mature repositories with stable conventions, especially Django and Sympy. &
\texttt{django\_\_django-11133}, \texttt{django\_\_django-17087},
\texttt{sympy\_\_sympy-19254}. &
The issue often hides exact expectations in tests or framework internals; topic
skills preserve these conventions across later tasks. \\
\midrule
TAU-Bench &
Retail exchanges/returns and airline reservation edits with multiple
constraints. &
\texttt{retail\_task\_5}, \texttt{retail\_task\_91},
\texttt{airline\_task\_13}. &
The state verifier rewards complete tool-call execution. Skills prevent
partial updates, wrong mutation tools, and guessed item or flight identifiers. \\
\midrule
CL-Bench &
Rule-application, procedural, and domain-knowledge contexts with local rubrics.
&
Context categories with many turnarounds include Rule System Application,
Procedural Task Execution, and Domain Knowledge Reasoning. &
Context skills behave like small local manuals: exact extraction rules, role
boundaries, output schemas, and calculation conventions. \\
\bottomrule
\end{tabular}
\end{table*}

\section{What the Harness Actually Learns}\label{app:learned-skills}

\begin{figure*}[t]
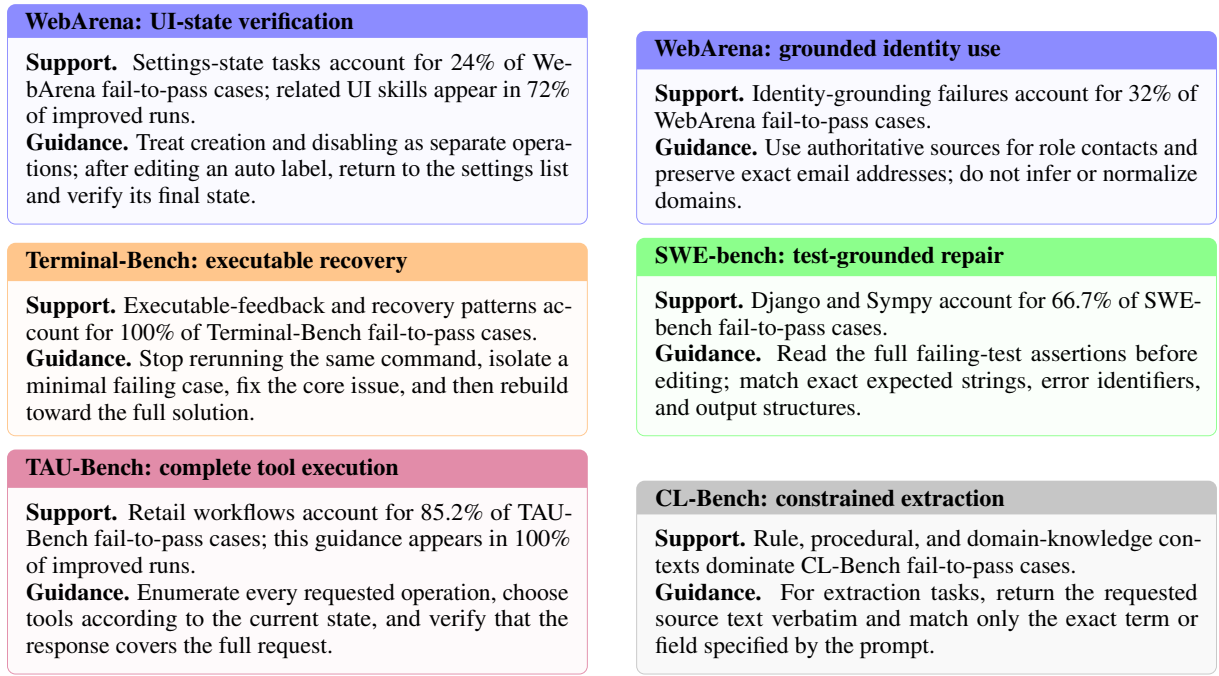

\centering
\begin{minipage}[t]{0.48\textwidth}
\begin{tcolorbox}[skillCard, colback=blue!2, colframe=blue!45,
title={WebArena: UI-state verification}]
\small
\textbf{Support.} Settings-state tasks account for 24\% of WebArena fail-to-pass cases; related UI skills appear in 72\% of improved runs.\\
\textbf{Guidance.} Treat creation and disabling as separate operations; after editing an auto label, return to the settings list and verify its final state.
\end{tcolorbox}
\end{minipage}
\hfill
\begin{minipage}[t]{0.48\textwidth}
\begin{tcolorbox}[skillCard, colback=blue!2, colframe=blue!45,
title={WebArena: grounded identity use}]
\small
\textbf{Support.} Identity-grounding failures account for 32\% of WebArena fail-to-pass cases.\\
\textbf{Guidance.} Use authoritative sources for role contacts and preserve exact email addresses; do not infer or normalize domains.
\end{tcolorbox}
\end{minipage}

\vspace{4pt}
\begin{minipage}[t]{0.48\textwidth}
\begin{tcolorbox}[skillCard, colback=orange!2, colframe=orange!45,
title={Terminal-Bench: executable recovery}]
\small
\textbf{Support.} Executable-feedback and recovery patterns account for 100\% of Terminal-Bench fail-to-pass cases.\\
\textbf{Guidance.} Stop rerunning the same command, isolate a minimal failing case, fix the core issue, and then rebuild toward the full solution.
\end{tcolorbox}
\end{minipage}
\hfill
\begin{minipage}[t]{0.48\textwidth}
\begin{tcolorbox}[skillCard, colback=green!2, colframe=green!45,
title={SWE-bench: test-grounded repair}]
\small
\textbf{Support.} Django and Sympy account for 66.7\% of SWE-bench fail-to-pass cases.\\
\textbf{Guidance.} Read the full failing-test assertions before editing; match exact expected strings, error identifiers, and output structures.
\end{tcolorbox}
\end{minipage}

\vspace{4pt}
\begin{minipage}[t]{0.48\textwidth}
\begin{tcolorbox}[skillCard, colback=purple!2, colframe=purple!45,
title={TAU-Bench: complete tool execution}]
\small
\textbf{Support.} Retail workflows account for 85.2\% of TAU-Bench fail-to-pass cases; this guidance appears in 100\% of improved runs.\\
\textbf{Guidance.} Enumerate every requested operation, choose tools according to the current state, and verify that the response covers the full request.
\end{tcolorbox}
\end{minipage}
\hfill
\begin{minipage}[t]{0.48\textwidth}
\begin{tcolorbox}[skillCard, colback=gray!5, colframe=gray!45,
title={CL-Bench: constrained extraction}]
\small
\textbf{Support.} Rule, procedural, and domain-knowledge contexts dominate CL-Bench fail-to-pass cases.\\
\textbf{Guidance.} For extraction tasks, return the requested source text verbatim and match only the exact term or field specified by the prompt.
\end{tcolorbox}
\end{minipage}

\caption{
Representative learned harness excerpts across benchmarks. Each card pairs a recurring fail-to-pass pattern with the corresponding evolved guidance. Percentages summarize matched baseline-fail/evolved-pass cases and contextualize the examples rather than establish single-skill causality.
}
\label{fig:learned-skill-excerpts}
\end{figure*}

Figure~\ref{fig:learned-skill-excerpts} shows real skill excerpts selected with
matched fail-to-pass support. The evidence line reports the relevant turnaround
family in the runs, while the rule line gives the core learned behavior.

\section{Prompt Templates}\label{app:prompts}

The prompts follow the same context-to-harness compilation logic across experiments: the solver reflects on an execution context to propose candidate lessons, and the evolver curates these candidates against the existing harness. Below, we provide representative prompt templates that capture this common proposal-and-curation process.

\subsection{Solver-Side Proposal Prompt}

\begin{tcolorbox}[breakable, colback=gray!5, colframe=gray!50, fontupper=\small,
title={\small Proposal prompt, compressed}]
\textbf{Inputs:} evaluation result, verifier details or rubric feedback,
trajectory signals, compressed trajectory, and related existing skills.\\[2pt]
\textbf{Analyze the execution.}
For each distinct issue, identify what went wrong and the concrete missing
action, command, tool call, navigation step, or domain rule.\\[2pt]
\textbf{Propose a reusable skill.}
Choose a broad topic, decide \texttt{NEW}, \texttt{ENHANCE}, or \texttt{NONE},
write a trigger-style description, and provide short bullet-point content with
techniques and gotchas.\\[2pt]
\textbf{Filter aggressively.}
Skip generic advice, basic tool usage, exact task replay, and skills that would
not help unseen tasks.
\end{tcolorbox}

\subsection{Evolver-Side Curator Prompt}

\begin{tcolorbox}[breakable, colback=gray!5, colframe=gray!50, fontupper=\small,
title={\small Curator prompt, compressed}]
\textbf{Inputs:} topic, current skill library, budget, and proposals from the
latest batch.\\[2pt]
For each proposal, choose \texttt{ACCEPT}, \texttt{MERGE}, or \texttt{SKIP}.
Prefer merging over duplication, respect the budget, require a clear trigger
description, keep content short and actionable, and apply a generalizability
test such as usefulness for multiple unseen tasks.
\end{tcolorbox}

\subsection{General Skill Curator}

\begin{tcolorbox}[breakable, colback=gray!5, colframe=gray!50, fontupper=\small,
title={\small General skill curator, compressed}]
Analyze repeated patterns across tasks rather than within a single topic.
Create or update a general skill only when a pattern appears across multiple
contexts. General skills must avoid context-specific references and encode
procedures that can guide planning, verification, recovery, or tool use across
tasks.
\end{tcolorbox}

\section{Experimental Details}
\label{app:details}
All experiments use the same open-source pipeline with benchmark-specific adapters. WebArena uses browser workers with persistent application state. Terminal-Bench uses task-specific Docker images and executable verifiers. SWE-bench uses isolated repository environments and unit-test evaluation. TAU-Bench uses the official tool-use environment and verifier. CL-Bench uses a rubric-based judge pipeline.

The harness is stored as Markdown skill files with lightweight YAML metadata, making the learned artifact inspectable and easy to transfer across runs. Each entry contains a trigger describing when it should be retrieved, a short actionable rule or procedure, optional evidence linking it to prior executions, and a scope indicating whether it is intended as cross-task guidance or task-type guidance. For harness selection, we use Claude Sonnet 4.5 across all experiments to retrieve relevant skills from the current harness before task execution.

Unless otherwise specified, we use a batch size of 16. We set the maximum number of general skills and the maximum number of skills under each task-type topic to 5. This budget encourages the evolver to merge overlapping guidance and avoid accumulating overly detailed or task-specific records.

\section{Potential Risks}
\algname{} evolves skill libraries that are injected into agent prompts as procedural guidance. The learned skills are designed to capture task-solving patterns such as navigation flows, tool-use procedures, verification steps, and recovery strategies. In our experiments, all benchmarks are run in sandboxed environments with no access to real user data, external accounts, or production systems. Therefore, the evaluated setting poses minimal practical risk beyond standard benchmark execution.

\section{Use Or Create Scientific Artifacts}
Our work uses five established public benchmarks: WebArena-Infinity, TerminalBench2, SWE-bench Lite, CL-Bench, and TAU-Bench. We do not modify the benchmarks themselves. We use foundation models (Claude and other models) via their standard AWS Bedrock APIs. We will release the \algname{} pipeline code and all evolved skill libraries upon publication.

\subsection{Cite Creators Of Artifacts}
All benchmarks and models are properly cited in the main text.

\subsection{Discuss The License For Artifacts}
All benchmark resources used in this work are publicly available under their respective research or open-source licenses. In particular, WebArena Infinity and TAU-Bench are released under MIT licenses, WebArena is released under Apache-2.0, and CL-Bench is released under a custom evaluation-only license. Our pipeline code will be released under the MIT License.

\subsection{Artifact Use Consistent With Intended Use}
All benchmarks are used for their intended purpose of evaluating agent capabilities. The foundation models are used via their standard APIs in accordance with their terms of service.

\subsection{Data Contains Personally Identifying Info Or Offensive Content}
The benchmarks use synthetic or sanitized scenarios. SWE-bench uses public GitHub issues. To our knowledge, none contain personally identifying information or offensive content.

\section{Computational Experiments}\label{sec:app-compute}

\subsection{Compute Budget}
The full experimental suite is computationally expensive because it covers multiple benchmarks, solver models, evolution settings, and ablation configurations. Across all main experiments and diagnostic runs, the total API cost was on the order of \(\$100\mathrm{K}\). For a single benchmark configuration, we typically run 16 parallel workers. Depending on benchmark complexity, task timeout, and environment overhead, one full run takes approximately 2--8 hours of wall-clock time. Web and tool-use benchmarks are usually faster, while software-engineering and command-line tasks tend to take longer because they require repository setup, Docker execution, test running, or verifier calls.

\subsection{Experimental Setup And Hyper-params}
Key hyperparameters are described in Appendix~\ref{app:details}. We use fixed random seeds (42) for task shuffling to ensure reproducibility.

\subsection{Descriptive Statistics}
We report pass rate (\%) as the primary metric across all benchmarks. For statistical reliability, we report results over the full task set for each benchmark rather than subsampling. The performance is averaged over three runs.

\subsection{Parameters For Packages}

The project primarily used AWS Bedrock API for foundation model inference, together with standard AWS infrastructure services such as EC2, S3, EBS/FSx, CloudWatch, IAM, and VPC. The software environment used Python 3.10+, boto3 1.34+, botocore 1.34+, PyTorch 2.0+, CUDA 12.0+, Docker, and Ubuntu 24.04.

\section{AI Assistants In Research Or Writing}

This paper studies LLMs as the research object and uses frozen LLM backbones
for experimental inference. These models are not used to create new benchmark
labels or to fabricate experimental results. Reported numerical results come
from the experimental code and are checked by the authors.

AI assistant tools were used for polishing manuscript. They were not used to
decide scientific claims, select reported results, or replace author
verification.

\end{document}